\documentclass[11pt]{article} 
\usepackage{rldmsubmit,palatino}
\usepackage{graphicx}
\usepackage{subcaption}
\usepackage{caption}
\usepackage{float}
\usepackage{amsmath}
\usepackage[sort,numbers]{natbib}
\usepackage{algpseudocode}
\usepackage{algorithm}
\usepackage{listings}
\usepackage{wrapfig}

\newcommand{\oparguments}{\overline{v}}
\newcommand{\preconditions}{P}
\newcommand{\addeffects}{E^+}
\newcommand{\deleteeffects}{E^-}

\newif\ifnotes
\notestrue 
\ifnotes
\newcommand{\nnote}[1]{\textcolor{magenta}{\textbf{Nishanth}: #1 }}
\newcommand{\anote}[1]{\textcolor{magenta}{\textbf{Annie}: #1 }}
\else
\newcommand{\nnote}[1]{}
\newcommand{\anote}[1]{}
\fi

\title{Guided Exploration for Efficient Relational Model Learning}

\author{
Annie Feng\thanks{Correspondence to feng.annie@alum.mit.edu} \\
MIT CSAIL\\
\texttt{feng.annie@alum.mit.edu} \\
\And
Nishanth Kumar \\
MIT CSAIL \\
\texttt{njk@csail.mit.edu} \\
\And
Tom\a'as Lozano-P\a'erez \\
MIT CSAIL \\
\texttt{tlp@csail.mit.edu} \\
\And
Leslie Pack-Kaelbling \\
MIT CSAIL \\
\texttt{lpk@csail.mit.edu} \\
\\
}

\pdfoutput=1
\begin{document}

\maketitle

\begin{abstract}
Efficient exploration is critical for learning relational models in large-scale environments with complex, long-horizon tasks. Random exploration methods often collect redundant or irrelevant data, limiting their ability to learn accurate relational models of the environment. Goal-literal babbling (GLIB) improves upon random exploration by setting and planning to novel goals, but its reliance on random actions and random novel goal selection limits its scalability to larger domains.
In this work, we identify the principles underlying efficient exploration in relational domains: (1) operator initialization with demonstrations that cover the distinct lifted effects necessary for planning and (2) refining preconditions to collect maximally informative transitions by selecting informative goal-action pairs and executing plans to them. To demonstrate these principles, we introduce Baking-Large, a challenging domain with extensive state-action spaces and long-horizon tasks. We evaluate methods using oracle-driven demonstrations for operator initialization and precondition-targeting guidance to efficiently gather critical transitions. Experiments show that both the oracle demonstrations and precondition-targeting oracle guidance significantly improve sample efficiency and generalization, paving the way for future methods to use these principles to efficiently learn accurate relational models in complex domains.

\end{abstract}



\startmain 
\section{Introduction}
\label{sec:intro}
A core challenge for autonomous agents is solving complex, long-horizon tasks --- such as having a web-based AI agent plan a user's travel itinerary, or having a robot bake a cake --- in large-scale environments.
Relational domain models provide a powerful framework for solving such tasks by enabling efficient planning over such state and action spaces. 
These models leverage symbolic predicate relations \cite{Fox2003} to represent states and actions compactly, allowing for generalization across similar problems. 
Additionally, the structure inherent to such models can provide leverage for efficient learning algorithms~\cite{zpk}.
In this work we are interested in the problem of learning relational models in large-scale domains from online interaction data in those domains.

Our core focus is on studying effective exploration strategies that enable efficient online learning.
We build on goal-literal babbling (GLIB) \cite{glib}, a previous approach that gathers informative training data through goal selection. GLIB proved to be substantially more effective than random exploration for solving relatively simple problems. However, in this paper, we show that GLIB's performance degrades significantly with even modest increases in the complexity of the planning problem. This suggests the need for improved exploration strategies in complex relational model-learning problems.

Intuitively, GLIB initially relies on taking random actions to collect transitions until a dataset can be built up such that some initial operators can be learned.
Once these are learned, it attempts to plan with these operators to try to reach novel goals.
If these plans are incorrect in that they cannot be successfully executed in the environment, then the agent will collect an unexpected transition when attempting to execute them, which will cause the operator learner to `fix' the learned operators.
In this fashion, the operators are refined over time. In this work, we extend this idea to more complex settings.




In this paper, we make progress toward—but do not fully achieve—the goal of designing a fully autonomous exploration strategy. Building on the foundation of GLIB, we extend the method with oracle-driven initialization and guidance. Specifically, we introduce a strategy that uses oracle information to: (1) initialize the agent’s learning with demonstrations that cover critical action outcomes, and (2) guide goal selection to maximize the informativeness of subsequent exploration.

Our results demonstrate that combining oracle-driven initialization and guidance with improved planning failure-handling enables significant improvements over GLIB. These ideas can serve to focus future research in complex and realistic relational model-learning problems in larger, longer-horizon planning domains.


\section{Problem Formulation and Background}
\label{sec:background}

\subsection{Environments and Tasks}
\label{subsec:envs}
An environment is defined as a tuple $\langle \mathcal{S}, \mathcal{A}, H, T, \mathcal{P} \rangle$.
Within an environment, a task is defined by a tuple $\langle I, \mathcal{O}, G \rangle$. 
Here, $\mathcal{O}$ is some finite set of objects and $\mathcal{P}$ is a finite set of predicates defined on those objects.
Each predicate is a binary classifier that takes in a sequence of objects and outputs a boolean (e.g. \texttt{OnTop(?object1, ?object2)}, \texttt{Cracked(?object)}). A predicate with specific objects substituted is called a \textit{ground atom} (e.g.  \texttt{OnTop(bowl1, table1)}, \texttt{Cracked(egg2)}). The (discrete) state space $\mathcal{S}$ of a task is defined by all possible values all ground atoms could take.
Each action is a skill that takes in particular objects and affects a change to the state (e.g. \texttt{MoveForward(robot}). 
$T(s' | s, a)$ is the deterministic transition model where $s, s' \in \mathcal{S}$ and $a \in \mathcal{A}$. 
Within a task, $I$ is an initial ground atom state, and $G$ is a goal condition expressed as a conjunction of a small set of ground atoms.
The goal $G$ is achieved in any state $s_{x}: G \subseteq s_{x}$.
Finally, $H$ is the maximum episode length (i.e., the maximum number of consecutive actions that can be taken) to solve any task.

We are primarily interested in learning to solve tasks within complex, large-scale environments.
As an illustrative example, consider `Baking-Large': a variant of the Baking domain introduced by~\citet{pddlgym}. In this domain, an agent can perform a variety of baking-related tasks, such as baking a cake or souffle, by collecting ingredients, mixing them together in containers, and using baking and cooking appliances.
Baking-Large has several characteristics that make exploration rather challenging. 
First, there are up to 2523 ground actions (action predicates applied to objects) in tasks of interest, which makes random exploration over the state space infeasible. Second, there are many state predicates (around 1600 ground atoms in our training tasks: about $2^{1600}$ potential states), but only a small subset of those states are achievable. Third, achieving any of the meaningful goals may require a very long sequence of actions; for example, baking two souffles can require over 26 prerequisite actions. Thus, exploration needs to be focused and cannot rely on random sampling in either the state or action space as GLIB does.

\subsection{Operator Learning and Planning}
\label{subsec:op-learning}
Our main interest is to learn from data to solve a variety of tasks within a particular environment.
Specifically, we learn a set of \textit{symbolic operators}.
Each operator is a tuple $\langle \oparguments, \preconditions, \addeffects, \deleteeffects, a \rangle$. 
Here, $\oparguments$ are variables representing the operator's arguments, $\preconditions$, $\addeffects$ and $\deleteeffects$ are sets of predicates representing operator preconditions, add effects and delete effects respectively, and $a$ is an action associated with the operator. 
Similar to predicates, operators can be \textit{ground} when specific objects are substituted into the typed variable arguments $\oparguments$.
Thus, the same operator can be used across multiple tasks with different objects within the same environment.

Learning operators requires collecting a transition dataset $\mathcal{D}$ via online interaction with tasks in the environment.
Specifically, $\mathcal{D}$ is a set of transition tuples $(s, a, s')$ representing a state $s$, action $a$, and next state $s'$ observed according to the transition model $T(s, a)$.
We leverage an existing approach~\citep{zpk} to perform operator learning given a particular dataset $\mathcal{D}_i$ to yield some set of operators $\Omega_{i}$. 

Given these operators, as well as the provided predicate set $\mathcal{P}$, we will leverage a symbolic planner to yield a sequence of ground operators (i.e., a \textit{plan}) that correspond to an action sequence that the agent can execute in a given task $\langle I, \mathcal{O}, G \rangle$ to get from $I$ to a state satisfying $g$.

\subsection{Exploration Problem Setup}
\label{subsec:exploration}
We assume a standard setting in which the agent is provided with a set of \textit{training tasks}.
Within each training task, it can explore for $H$ steps before the environment is randomly reset to a different task from the training task set; each of these is considered a training episode.
All transitions experienced by the agent accumulate in a dataset $\mathcal{D}$ and are fed to the operator learner to constantly update the operator set $\Omega$.
For the purposes of evaluation, we ask the agent to attempt to solve a set of \textit{testing tasks} using operators $\Omega$ and predicates $P$. The testing tasks evaluate single operators and the synergy of all the learned operators on the longest horizon tasks possible in the domain. For example, in Baking-Large, there are 19 tasks to evaluate each of the 19 operators and 3 long-horizon tasks with minimum solution lengths of about 22.


The agent's objective is to collect the smallest dataset possible such that the corresponding operator set solves all the testing tasks. 
When it is exploring, the agent will behave according to an \textit{exploration policy} $\pi_{\text{explore}}: S \rightarrow A$ that is a mapping from environment states to actions. When it is evaluated on testing tasks, the agent will behave according to a \textit{learned policy} $\pi_{\text{agent}}: S \rightarrow A$ informed by the data collected through exploration. These "implicit" policies are obtained by planning using a symbolic planner that takes as input the transition model, an initial state, and a goal state to output plans. 
Additionally, the oracle will respond to agent queries via its own policy $\pi_{\text{guidance}}$, which is a mapping from the agent's learned operators and a task to goals, and provide plans to those goals using its implicit policy $\pi_{\text{oracle-plan}}$, which is a mapping from environment states to actions.

\subsection{Exploration via Goal-Literal Babbling}
\label{subsec:glib-background}
We build on GLIB~\citep{glib}: an approach that explores by `babbling' goals and then leveraging planning with the current operator set to attempt to reach these goals.
GLIB's exploration policy ($\pi_{\text{explore}}$) operates by randomly sampling a previously-unvisited goal (depending on the GLIB variant, the goal can be a lifted or ground conjunction of a specified number of predicates) from the given predicates and objects, and then attempting to plan to it using the current learned operator set $\Omega$.
If a plan is found, the agent attempts to execute it until it completes or fails. 
If no plan is found after $N$ attempts, GLIB defaults to randomly selecting an action from the action space.
The authors also introduce an upper bound on GLIB called Oracle-BFS that leverages knowledge of a set of ground-truth operators for the environment as part of its exploration policy.
Specifically, Oracle-BFS's exploration policy is to randomly select a learned operator $\omega$ from the current set $\Omega$ such that the predicted effects under the ground truth and learned models do not match. If all match, Oracle-BFS performs a modified breadth-first search (expanding a maximum of 50 random neighbors and limiting the search depth to 2 nodes) in the models, checking for any future mismatches, and falls back to random actions when none are found. 
 
\section{Guiding Principles for Efficient Exploration}
\label{sec:method}

Efficient exploration in complex planning problems requires a clear understanding of the data needed for relational model learning and the strategies to acquire it. In this section, we describe the guiding principles for gathering training data effectively in harder planning problems such as those encountered within Baking-Large.

\subsection{What Data Is Needed?}
GLIB initializes operators through random actions and refines preconditions by selecting random novel goals. From this perspective, we identify the types of data required for a minimal and sufficient dataset:

\begin{enumerate}
    \item \textbf{Type 1: } Transitions with a previously unseen set of lifted (i.e., not ground) effects that define a new operator. 
    \item \textbf{Type 2: } Transitions that provide counterexamples to an existing operator’s lifted preconditions. 
\end{enumerate}

 GLIB operates by first collecting a Type 1 transition to initialize the operator. Once this is identified, then GLIB collects Type 2 transitions to refine preconditions of the operator created from the Type 1 transition. 
GLIB gathers Type 1 transitions primarily through random actions—either by executing purely random actions when no operators have been learned or by planning to and achieving a random goal before executing a random action. It collects Type 2 transitions by planning to random goals and executing plans that fail, resulting in unexpected effects. These failed plan executions are critical because they directly reveal inaccuracies in the learned preconditions, leading to their improvement.

We aim to improve upon GLIB by collecting Type 1 and then Type 2 transitions via more focused methods than having to rely on random sampling of actions or goals.

\subsection{Algorithm Design for Gathering Training Data With An Oracle}
GLIB's reliance on random goal and action selection often leads to redundancy and inefficiency, especially in large action and state spaces. To address these challenges, we propose principles to make the underlying strategy more effective--- initializing operators and refining their preconditions---and consider the ideal case with initialization and precondition-targeting guidance from an oracle.

\subsubsection{Initializing Operators}
In the ideal case, initialization would involve access to all critical lifted action effects necessary for solving planning problems. Each operator could then be initialized using a single transition datapoint. The feasibility and acquisition of such demonstrations depend on the domain; for instance, they might be provided by a large language model or a human expert. These demonstrations ensure that the operator set includes all distinct lifted effects required for planning.

This allows us to learn an initial set of operators that correspond to the ground-truth operators by their lifted effects, enabling precondition comparisons and refinements.

\subsubsection{Refining Preconditions}

To refine initialized operators’ preconditions, the transitions must be chosen to maximize informativeness for the learning algorithm. Informative transitions are characterized by plan execution failures where unexpected effects occur. We categorize these transitions into three cases based on the relationship between the learned preconditions and the ground-truth preconditions:

\begin{itemize}
    \item \textbf{Strictly Stronger Preconditions:} The learned preconditions are stronger than the ground truth. Goals are selected to satisfy the ground-truth preconditions while violating the stronger components of the learned preconditions.
    
    \item \textbf{Strictly Weaker Preconditions:} The learned preconditions are weaker than the ground truth. Goals are selected to satisfy the learned preconditions while violating the missing components of the ground-truth preconditions. 
    
    \item \textbf{Mixed Preconditions:} The learned preconditions have both stronger and weaker components. Goals are selected to either (1) satisfy the learned preconditions and violate the missing weaker components, or (2) satisfy the ground-truth preconditions and violate the stronger components.    
\end{itemize}

In precondition-targeting oracle guidance, the oracle randomly selects a learned operator with incorrect preconditions. Out of all possible informative goals for the operator, the oracle prioritizes sampling novel goals with the highest \textit{dissonance}—measured as the number of literals that are stronger or weaker compared to the true preconditions. The oracle makes a plan to each goal in this order until a plan is found to one of the goals. After achieving the goal, the agent grounds the associated lifted learned operator and executes it: this final action elicits the key transition that improves the learned model. Finally, once the plan finishes or the plan fails in execution, then the environment is reset to the starting state.

\section{Experiments}
We study how these principles enable the learning of generalizable theories in our experiments. Specifically, we analyze how oracle demonstrations and precondition-targeting oracle guidance affect performance across multiple domains. The methods evaluated include:

\begin{itemize}
   \item \textbf{GLIB\_L2}: No oracle guidance or initialization. Performs standard GLIB exploration with random novel goals and random actions. Goals are conjunctions of two lifted literals.
      \item \textbf{Oracle-BFS}: Modified BFS for mismatches in the predicted effects under the learned and ground-truth models (BFS oracle guidance) and no oracle initialization. An upper bound on GLIB defined by GLIB authors.
    \item \textbf{Oracle-Precondition-Targeting-Demos}: Incorporates our oracle initialization and and precondition-targeting oracle guidance using the ground-truth model. 
    \item \textbf{GLIB\_L2-Demos}: No oracle guidance and oracle initialization. GLIB\_L2 initialized with the oracle demonstrations.
   \item \textbf{Oracle-BFS-Demos}: BFS oracle guidance and oracle initialization.
\end{itemize}

\begin{figure}[H]
    \centering
    \begin{subfigure}{0.22\textwidth}
        \centering
        \includegraphics[width=\textwidth]{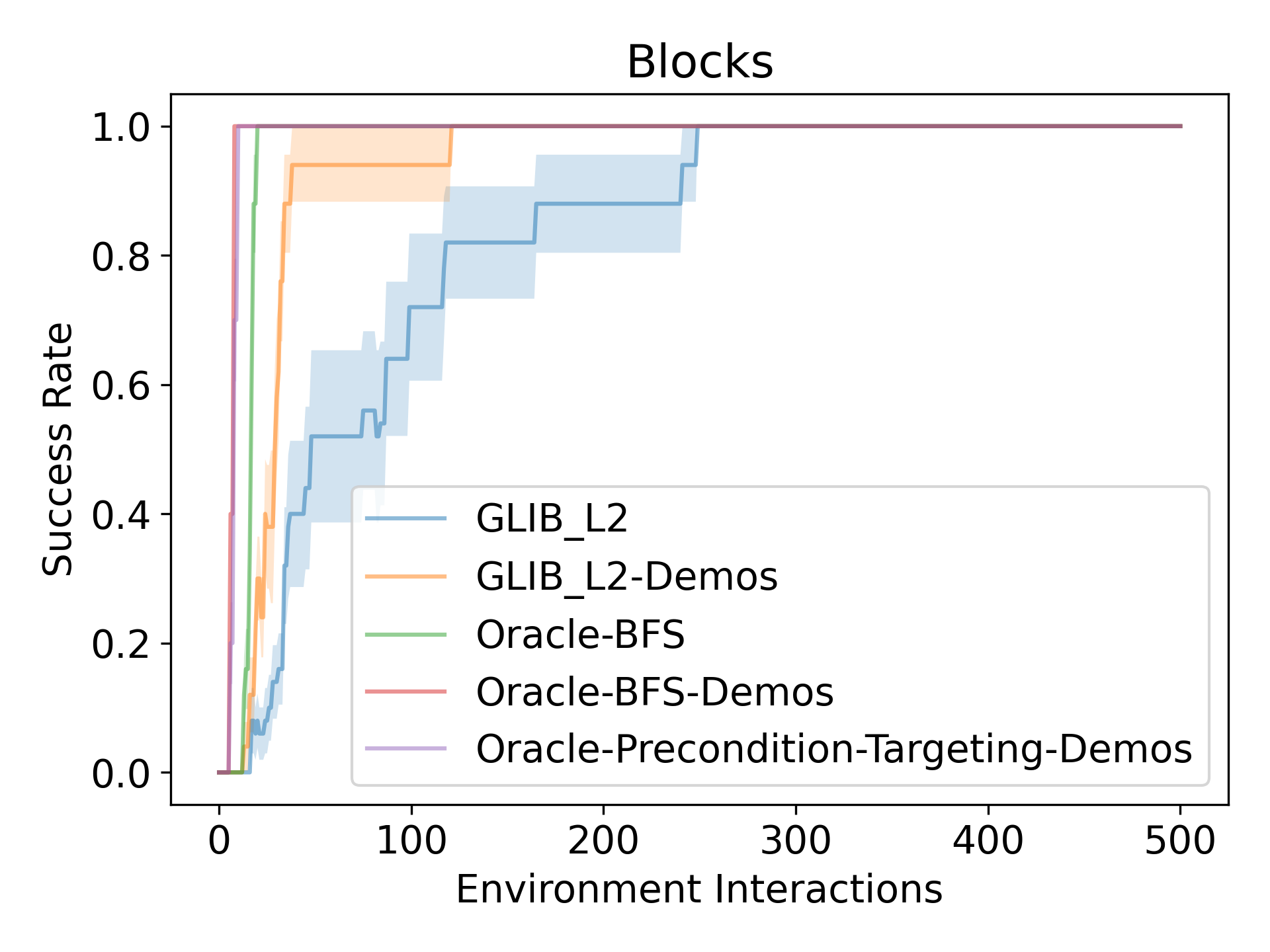}
    \end{subfigure}
    \begin{subfigure}{0.22\textwidth}
        \centering
        \includegraphics[width=\textwidth]{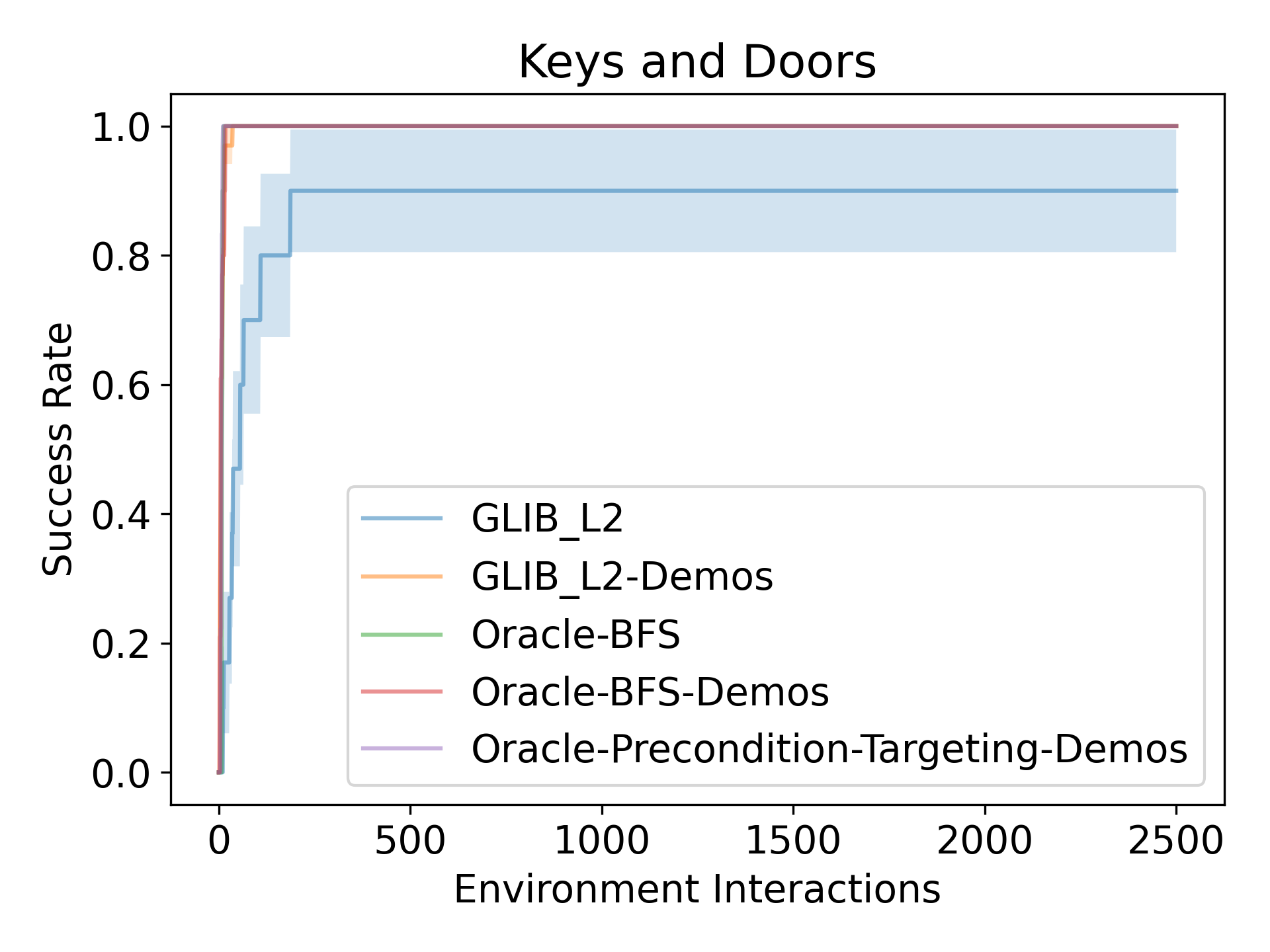}
    \end{subfigure}
    \begin{subfigure}{0.22\textwidth}
        \centering
        \includegraphics[width=\textwidth]{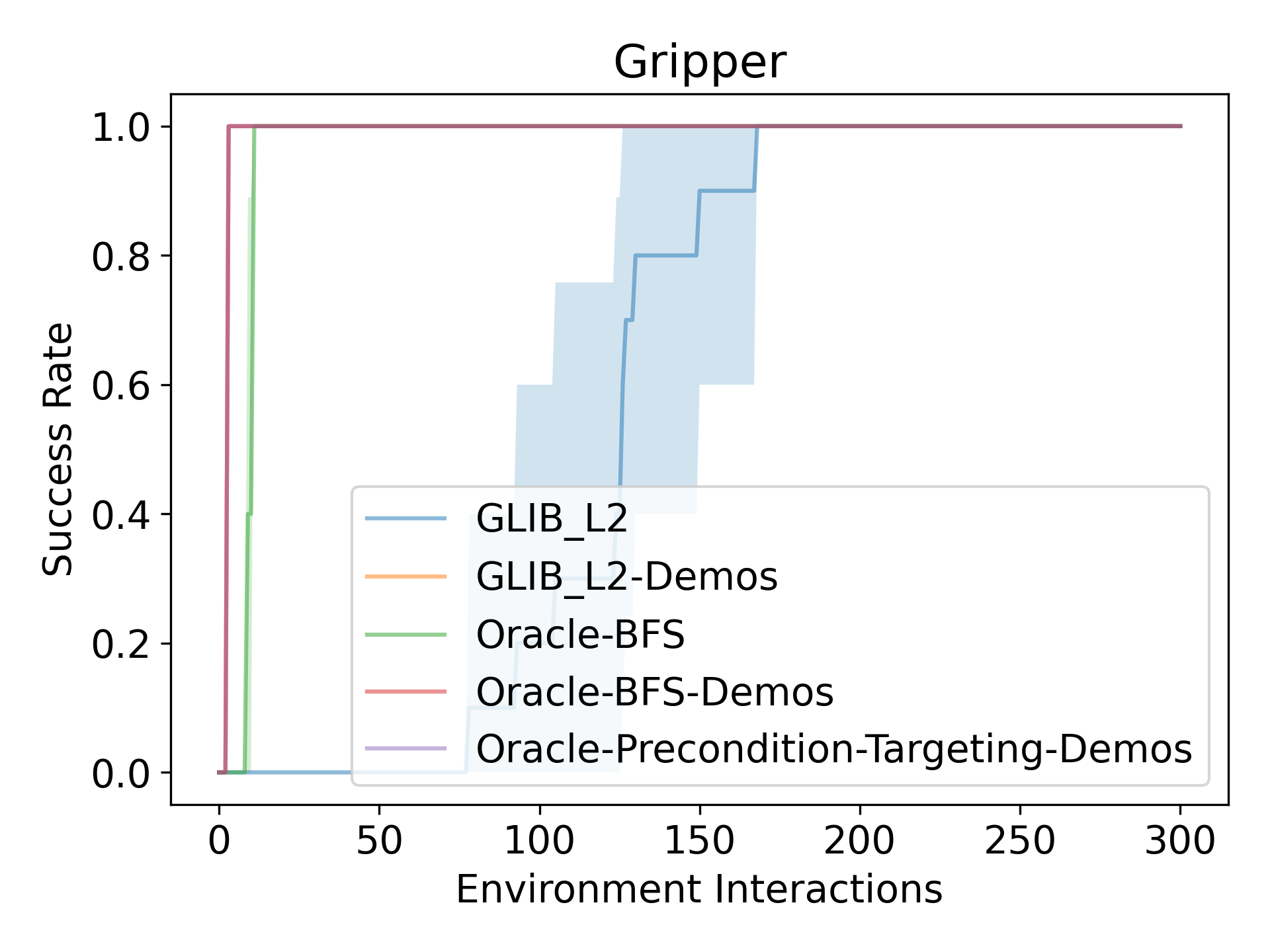}
    \end{subfigure}
    \begin{subfigure}{0.22\textwidth}
        \centering
            \includegraphics[width=\textwidth]{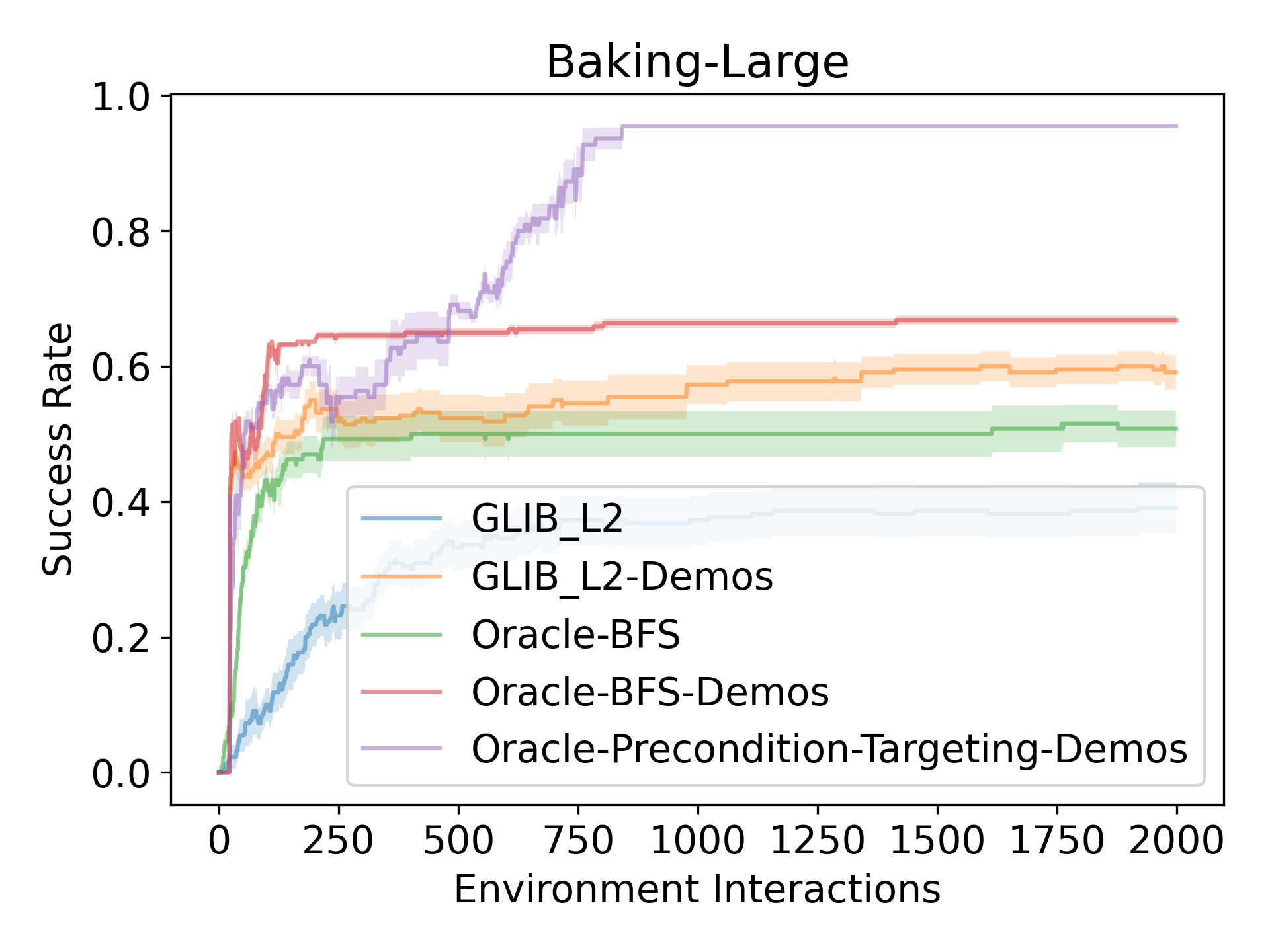}
    \end{subfigure}
    \caption{Success rate vs. number of environment interactions plots for all our approaches on all our domains. We run 10 independent random seeds and report the mean (solid line), as well as standard error (shading).}
    \label{fig:row2}
\end{figure}

Our results reveal that oracle demonstrations substantially improve performance in all domains. In fact, oracle demonstrations alone are sufficient to completely learn  Gripper (only 3 transitions are required) but are insufficient for the other domains.
Additionally, precondition-targeting oracle guidance significantly improves performance in complex domains like Baking-Large by refining operator preconditions. Only in the very simple domains, Oracle-BFS performs just as well as Oracle-Precondition-Targeting-Demos.

Oracle-BFS struggles in Baking-Large due to limitations in the modified breadth-first-search. In Baking-Large, the large number of ground actions makes it infeasible to traverse the graph of plans through modified BFS. The modified BFS fails to identify predicted effects that differ between the learned model and the ground-truth model in Baking-Large, so Oracle-BFS defaults to random actions, which are ineffective in the large domain. Even with oracle demonstrations, Oracle-BFS-Demos has the same problem. In contrast, Oracle-Precondition-Targeting-Demos consistently learns generalizable theories by planning to collect informative transitions.

These results show that our initialization and precondition refinement principles allow us to learn highly generalizable theories much more efficiently than GLIB or Oracle-BFS.

\section{Conclusion and Future Work}
In this work, we outlined the principles underlying GLIB exploration, identifying the types of training data needed for efficient relational model learning and proposing principles to acquire this data efficiently. We introduced Baking-Large, a challenging domain that highlights the limitations of prior approaches like GLIB and Oracle-BFS, and demonstrated how our principles enable learning in such domains.
Our results show that oracle demonstrations alone significantly improve performance but are not enough for learning complex operators. To address this, we showed how precondition-targeting oracle guidance---achieving maximally dissonant preconditions and executing these operators---helps to completely learn the operators.

Future work should focus on designing approximate methods that emulate our approach but are feasible for real-world applications. This includes leveraging large language models for generating initial demonstrations and guiding goal selection, as well as developing practical frameworks for human-robot teaching scenarios where the ground-truth model is unavailable. By operationalizing these principles, we can move closer to enabling efficient and autonomous learning in real-world relational domains.

\bibliographystyle{unsrtnat}
\bibliography{main}

\begin{thebibliography}{4}
\providecommand{\natexlab}[1]{#1}
\providecommand{\url}[1]{\texttt{#1}}
\expandafter\ifx\csname urlstyle\endcsname\relax
  \providecommand{\doi}[1]{doi: #1}\else
  \providecommand{\doi}{doi: \begingroup \urlstyle{rm}\Url}\fi

\bibitem[Fox and Long(2003)]{Fox2003}
M.~Fox and D.~Long.
\newblock Pddl2.1: An extension to pddl for expressing temporal planning
  domains.
\newblock \emph{Journal of Artificial Intelligence Research}, 20, 2003.
\newblock URL \url{http://dx.doi.org/10.1613/jair.1129}.

\bibitem[Kaelbling et~al.(2011)Kaelbling, Pasula, and Zettlemoyer]{zpk}
Leslie Kaelbling, Hanna Pasula, and Luke Zettlemoyer.
\newblock Learning symbolic models of stochastic domains.
\newblock \emph{Journal of Artificial Intelligence Research}, 29, 10 2011.
\newblock \doi{10.1613/jair.2113}.

\bibitem[Chitnis et~al.(2020)Chitnis, Silver, Tenenbaum, Kaelbling, and
  Lozano{-}P{\'{e}}rez]{glib}
Rohan Chitnis, Tom Silver, Joshua~B. Tenenbaum, Leslie~Pack Kaelbling, and
  Tom{\'{a}}s Lozano{-}P{\'{e}}rez.
\newblock {GLIB:} exploration via goal-literal babbling for lifted operator
  learning.
\newblock \emph{CoRR}, abs/2001.08299, 2020.
\newblock URL \url{https://arxiv.org/abs/2001.08299}.

\bibitem[Silver and Chitnis(2020)]{pddlgym}
Tom Silver and Rohan Chitnis.
\newblock Pddlgym: Gym environments from pddl problems, 2020.
\newblock URL \url{https://arxiv.org/abs/2002.06432}.

\end{thebibliography}

\end{document}